\documentclass{article} % For LaTeX2e
\usepackage{iclr2021_conference,times}

\usepackage{amsmath,amsfonts,bm}

\def\eqref#1{equation~\ref{#1}}
\def\1{\bm{1}}

\DeclareMathAlphabet{\mathsfit}{\encodingdefault}{\sfdefault}{m}{sl}
\SetMathAlphabet{\mathsfit}{bold}{\encodingdefault}{\sfdefault}{bx}{n}

\DeclareMathOperator*{\argmax}{arg\,max}

\usepackage{hyperref}
\usepackage{url}
\usepackage{tcolorbox}
\usepackage{enumitem}

\title{Scaffolding Simulations with Deep Learning \\ for High-dimensional Deconvolution}

\author{Anders Andreassen\thanks{Author list ordered alphabetically.} \\
Google\\
\small{\texttt{ajandreassen@google.com}}
\And
Patrick T. Komiske \\
MIT, IAIFI \\
\small{\texttt{pkomiske@mit.edu} }
\And
Eric M. Metodiev \\
MIT, IAIFI \\
\small{\texttt{eric.metodiev@gmail.com}}
\And
\AND
Benjamin Nachman\\
Berkeley Lab, BIDS \\
\small{\texttt{bpnachman@lbl.gov}}
\And
Adi Suresh\\
UC Berkeley \\
\small{\texttt{adisurtya@berkeley.edu}}
\And
Jesse Thaler\\
MIT, IAIFI \\
\small{\texttt{jthaler@mit.edu}}
}

\iclrfinalcopy % Uncomment for camera-ready version, but NOT for submission.
\begin{document}

\maketitle

\begin{abstract}
A common setting for scientific inference is the ability to sample from a high-fidelity forward model (simulation) without having an explicit probability density of the data.    We propose a simulation-based maximum likelihood deconvolution approach in this setting called \textsc{OmniFold}.  Deep learning enables this approach to be naturally unbinned and (variable-, and) high-dimensional.  In contrast to model parameter estimation, the goal of deconvolution is to remove detector distortions in order to enable a variety of down-stream inference tasks.  Our approach is the deep learning generalization of the common Richardson-Lucy approach that is also called Iterative Bayesian Unfolding in particle physics.  We show how \textsc{OmniFold} can not only remove detector distortions, but it can also account for noise processes and acceptance effects.
\end{abstract}

\section{Introduction}

In many scientific applications, an important step in data analysis is the mitigation of detector distortions.   When this is done at the level of an individual datum, this process is known as \textit{calibration} while when it is done at the level of an entire dataset, it is known as \textit{deconvolution}.   If detector distortions are known analytically, then one can perform maximum likelihood estimation to remove detector effects: $\hat{x}_T=\argmax_{x_G} p_{S|G}(x_D|x_G)$, where $T, D, S$, and $G$ stand for truth, data, simulation, and generation, respectively.  The data ($D$) are observed in a detector and we want to infer the pre-detector truth ($T$).  Our model of this process uses synthetic pre-detector generation ($G$) and post-detector simulation ($S$).  We often do not know $p_{S|G}$ analytically, especially when each data point is in a high- (and possibly variable-) dimensional space.  However, we often do have detailed simulations of our experimental devices and can sample from $p_{S|G}$.

Traditional deconvolution methods convert the problem of estimating $\hat{x}_T$ into a linear algebra problem.  If the data features are discretized into a finite number of histogram bins, then the entire dataset $x=\{x_1,...,x_n\}$ can be represented as a vector where the entries are the histogram bin contents.  The conditional density $p_{S|G}$ is then a matrix and can be estimated numerically by sampling.  A variety of regularized matrix inversion methods have been proposed and one of the most popular methods across scientific domains is the Richardson-Lucy (RL)~(\cite{Richardson:72,1974AJ.....79..745L}) approach that is also called Iterative Bayesian Unfolding~(\cite{DAgostini:1994fjx}) in particle physics.  The RL algorithm is an iterative approach where at each step, $p_{G|S}$ is estimated from $p_{S|G}$ and the estimate of $\hat{x}_T$ from the previous step, and then this matrix is applied to $x_\text{D}$ to arrive at a new estimate of $\hat{x}_T$.  This approach can be phrased in terms of an expectation-maximization (EM) algorithm and correspondingly proven to converge to the maximum likelihood estimate~(\cite{shepp1982maximum}).

While RL is a widely used algorithm, it has three key challenges.   First, it requires the data to be represented in a fixed number of bins.   This discretization is fixed at the beginning of the analysis and cannot be changed later (except to make it coarser).   Second, it is often impractical to add more dimensions and so typically strong dimensional reduction proceeds the deconvolution.  Third, because the number of dimensions are limited, the deconvolution cannot benefit from all possible auxiliary features that control the detector distortions.   

Deep learning has the potential to solve all three of these challenges.  We will introduce the \textsc{OmniFold} algorithm~(\cite{Andreassen:2019cjw}), which is an iterative approach based on likelihood-ratio estimation with deep learning classifiers.  A variety of other machine learning approaches have also been proposed, but they cannot process unbinned data~(\cite{Gagunashvili:2010zw,Glazov:2017vni,Datta:2018mwd,bunse2018unification}) or do not have desirable statistical properties (e.g.\ prior independence via maximum likelihood estimation)~(\cite{Bellagente:2019uyp,Bellagente:2020piv}).  This paper is organized as follows.  Section~\ref{sec:algorithm} introduces the \textsc{OmniFold} algorithm and extends the approach beyond Ref.~(\cite{Andreassen:2019cjw}) by accounting for noise processes and acceptance effects.    Numerical examples are presented in Sec.~\ref{sec:examples} and the paper ends with brief conclusions and outlook in Sec.~\ref{sec:conclusions}.

\section{The OmniFold Algorithm}
\label{sec:algorithm}

A complete deconvolution algorithm must be able to account for four effects:

\begin{description}
	\item[Part (1)] Noise processes.  In many cases, the data is a coherent superposition of a signal process and a background process.  This background process must be statistically subtracted before removing detector distortions.
	\item[Part (2)] Detector acceptance.  The detector elements may not capture all signal process examples due to finite thresholds and other acceptance effects.  
	\item[Part (3)] Detector distortions.  This is the classical convolution of the data with a noise function that must be statistically removed.
	\item[Part (4)] Detector efficiency.  Sometimes there are also truth-level thresholds so that some synthetic signal examples register synthetic detector signals but are not recorded as synthetic generated examples.
\end{description}

Binned approaches like RL can account for the above affects via 

\begin{align}
\hat{x}_T=p_{S|G}^{-1}\left((x_D-x_N)\odot \text{Pr}_{S\leftarrow G}\right)\odot \text{Pr}_{G\rightarrow S}^{-1}\,,
\end{align}

 where $\odot$ is the Hadamard (component-wise) product, $p_{S|G}^{-1}$ represents regularized matrix inversion (Part (3)), $x_N$ is a binned estimate of the noise process to be removed (Part (1)), $\text{Pr}_{S\leftarrow G}$ represents the fraction of simulation-level examples that have a corresponding generator-level example (Part (2)), and $\text{Pr}_{G\rightarrow S}$ represents the fraction of generator-level examples that have a corresponding simulation-level example (Part (4)).  The vector inverse $\text{Pr}_{G\rightarrow S}$ is computed component-wise.
 
 The \textsc{OmniFold} method achieves Parts (1)--(4) without binning using a series of binary, weighted classification tasks.  In particular, let $x_{D,i}$ be examples from data, $(x_{G,i},x_{S,i})$ be synthetic examples of the signal process, and $x_{N,i}$ are noise examples.  For the signal process, if one of $x_{G,i}$ or $x_{S,i}$ does not exist (acceptance or efficiency effects), it is set to a fixed dummy value $\emptyset$.  It is sometimes the case that the examples $(x_{G,i},x_{S,i})$ and $x_{N,i}$ represent more or fewer than one real event from data and so they come with weights $w_{\text{synth},i}$ and $w_{\text{noise},i}$, respectively. Let $g$ be parameterized as a deep neural network.  Then define,

  \begin{align}
  \label{eq:crossentropy}
 f(A,W_A;B,W_B) = \argmax_{g}\left( \sum_{a\in A} w_a\log g(a)+ \sum_{b\in B} w_b\log(1-g(b))\right)\,,
 \end{align}
 
 where $A,B,W_A,W_B$ are sets and $W_A, W_B$ correspond to sets of weights for the examples in $A$ and $B$, respectively.  In practice, the optimization in Eq.~\ref{eq:crossentropy} involves the usual regularization from machine learning training to avoid overfitting.  When the weights $w$ are all unity, then the neural networks $f$ are trained using the standard binary cross entropy loss functions.  Finally, define $\tilde{f}(A,W_A;B,W_B)=f/(1-f)$.  It is well-known~(\cite{hastie01statisticallearning,sugiyama_suzuki_kanamori_2012}) that when $f$ is sufficiently flexible and well-trained, $\tilde{f}$ approximates the weighted likelihood ratio of $x$ given $A$ or $B$.  Any other loss function that results in a (known) monotonic rescaling of the likelihood ratio would also work instead of Eq.~\ref{eq:crossentropy}.

 With all of these components, the \textsc{OmniFold} protocol is as follows:
 
 \begin{tcolorbox}
 \textbf{\textsc{OmniFold}:} \textit{(extending Ref.~(\cite{Andreassen:2019cjw}) by including Parts (1), (2), (4))}

 \begin{enumerate}[wide, labelwidth=!, labelindent=0pt,label=(\roman*)]
 \item $w_{D,i}=\tilde{f}(\{x_{D,i}\}\cup\{x_{N,i}\},\{1,...,1\}\cup\{-w_{\text{noise},i}\};\{x_{D,i}\},\{1,...,1\})(x_{D,i})$
 \item For $k$ in $N_\text{iterations}$ do:
 	\begin{enumerate}[wide, labelwidth=!, labelindent=10pt]
		\item $w_{\text{Step I},i}=\tilde{f}(\{x_{S,i}|x_{S,i}\neq \emptyset\},\{w_{\text{synth},i}|x_{S,i}\neq \emptyset\};\{x_{D,i}\},\{w_{D,i}\})(x_{S,i})$
		\item $\tilde{f}_\text{miss,I}=\tilde{f}(\{x_{G,i}|x_{S,i}\neq \emptyset\},\{w_{\text{Step I},i}|x_{S,i}\neq \emptyset\};\{x_{G,i}|x_{S,i}\neq \emptyset\},\{1,...,1\})$
		\item $w_{\text{Pull},i}=w_{\text{Step I},i}$ if $x_{S,i}\neq \emptyset$ and $w_{\text{Pull},i}=\tilde{f}_\text{miss,I}(x_{G,i})$ otherwise
		\item $w_{\text{Step II},i}=\tilde{f}(\{x_{G,i}\},\{w_{\text{Pull},i}\};\{x_{G,i}\},\{w_{\text{synth},i}\})(x_{G,i})$
		\item $\tilde{f}_\text{miss,II}=\tilde{f}(\{x_{S,i}|x_{G,i}\neq \emptyset\},\{w_{\text{Step II},i}|x_{G,i}\neq \emptyset\};\{x_{S,i}|x_{G,i}\neq \emptyset\},\{1,...,1\})$
		\item $w_{\text{Push},i}=w_{\text{Step II},i}$ if $x_{G,i}\neq \emptyset$ and $w_{\text{Push},i}=\tilde{f}_\text{miss,II}(x_{S,i})$ otherwise
		\item $w_{\text{synth},i}=w_{\text{Push},i}$
	\end{enumerate}
 \end{enumerate}
 
 The final result is the set $\{x_{G,i}|x_{G,i}\neq \emptyset\}$ with weights $\{w_{\text{synth},i}|x_{G,i}\neq \emptyset\}$, from which one can compute empirical probability densities, expectation values, etc. of weighted statistics. 
 
 \end{tcolorbox}
 
The intuition for the various steps are as follows.  \textsc{OmniFold} (i) achieves Part (1) through positive reweighting~(\cite{Nachman:2020fff}).  The core parts of the iterative process are (ii), (a), (c), (d), and (f).  Ref.~(\cite{Andreassen:2019cjw}) proved that these steps can be modeled as an EM-type algorithm and achieve the maximum likelihood estimate.  Step I weights ensure that modified detector-level simulation are statistically identical to the measured data.  The pairing between the pre- and post-detector data in simulation allow the Step I weights to be associated (`pulled') to the pre-detector features.  These weights are not sufficient because they are not a proper function of the pre-detector features since the same example can be mapped to different features through the stochastic mapping of the detector.  This requires the Step II weighting, which produces a proper function of the pre-detector features.  These weights can be associated (`pushed') with the post-detector features and the entire process can be repeated.  \textsc{OmniFold} (ii), (b) and (e) are required to handle acceptance and efficiency effects, respectively.  For examples that are missing pre- or post-detector features, we use positive reweighting~(\cite{Nachman:2020fff}) again to assign the average weight. An alternative choice would be to assign a weight of unity, which would mean that the prior is used for these examples.  When there is no noise process and no efficiency/acceptance effects, (i), (ii), (b), and (ii), (e) can be eliminated.  Finally, note that when the data $x_i$ are binned, \textsc{OmniFold} approximates the RL algorithm step by step.

\section{Numerical Examples}
\label{sec:examples}

To illustrate all of the deconvolution parts and the various \textsc{OmniFold} steps, we simulate a one-dimensional Gaussian.  In particular, $X_T\sim\mathcal{N}(0.2,0.8)$, $X_G\sim\mathcal{N}(0,1)$, and $X_D=X_T+Z, X_S=X_G+Z$ where $Z\sim\mathcal{N}(0,0.5)$.  There is a background noise process that is distributed as $\mathcal{N}(0,1.2)$ that occurs in 10\% of the measured examples.  Furthermore, 10\% of the measured values have no corresponding truth value and vice versa.  There are $10^5$ data and simulation examples.

The neural networks are all implemented in \textsc{TensorFlow} 1.15.0~(\cite{tensorflow}) and optimized using \textsc{Adam}~(\cite{adam}) with default values unless otherwise specified.  These networks are fully connected with three hidden layers with 50 nodes each and the rectified linear unit between hidden layers and a sigmoid as the output.  The networks were trained for 200 epochs with a batch size of 2000 and early stopping using a patience of 10 epochs.

\begin{figure}
    \centering
    \includegraphics[width=0.95\textwidth]{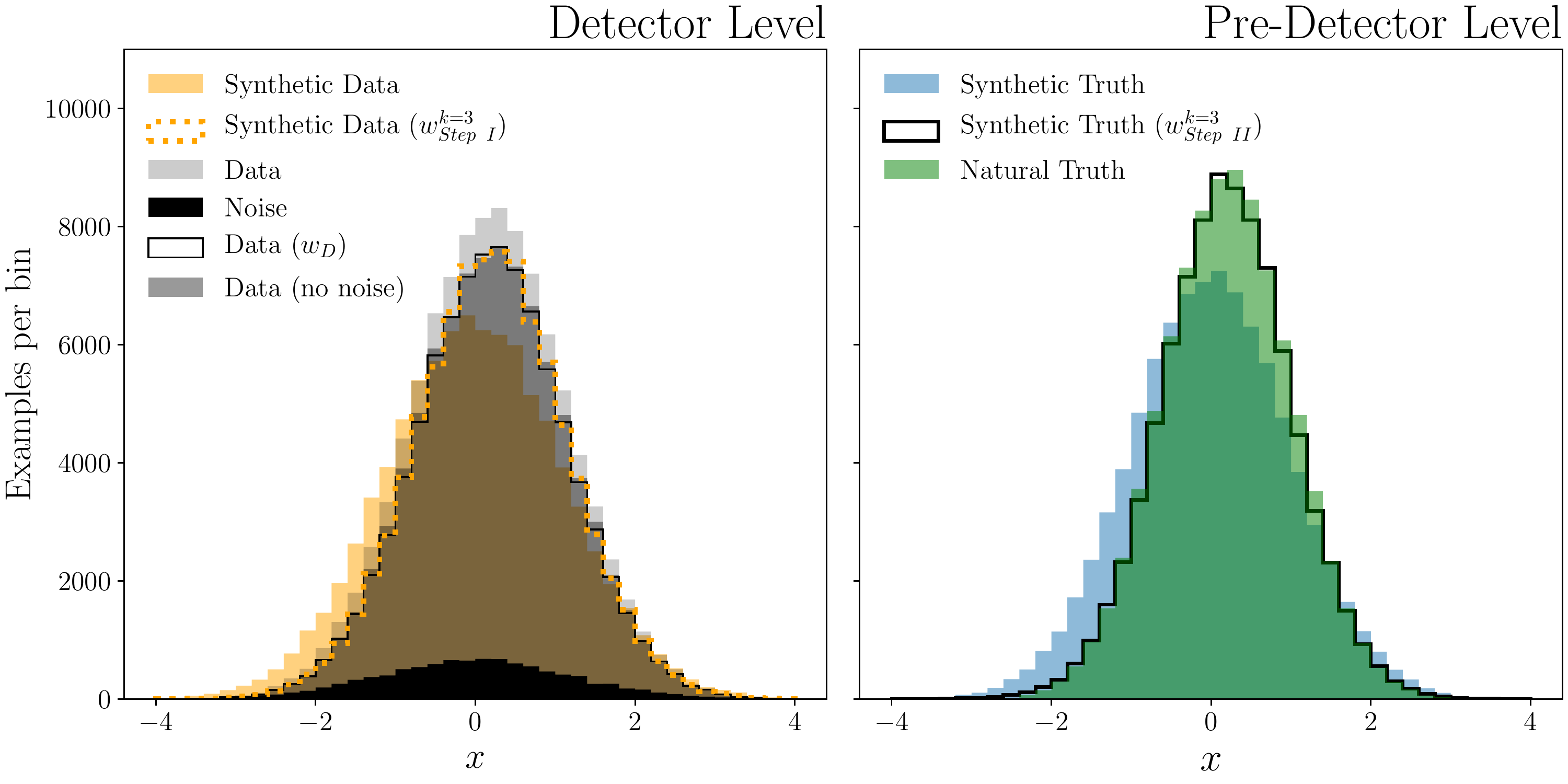}
    \caption{The complete \textsc{OmniFold} protocol for the one-dimensional Gaussian example.  Histograms on the left (right) are shown after (before) detector distortions.  The final result after three iterations is indicated with the weights $w_\text{Step I}^{k=3}$ on the left and $w_\text{Step II}^{k=3}$ on the right.}
    \label{fig:result}
\end{figure}

Figure~\ref{fig:result} presents the results of \textsc{OmniFold} with three iterations.  After three iterations, the results are nearly unchanged.  The unfilled histogram in the left plot of Fig.~\ref{fig:result} shows \textsc{OmniFold} (i), which agrees well with the noiseless data (dark grey).  This comparison would not be possible in practice as the noiseless data are not available.  The outcome of \textsc{OmniFold} (ii) is shown at detector-level on the left and prior to detector-effects on the right.  The orange dashed line (detector-level) agrees well with the black unfilled histogram (left) and the black unfilled histogram on the right agrees well with the green filled histogram.  The data are binned for illustration, but the result is naturally unbinned.

\textsc{OmniFold} can readily process multidimensional data.  Table~\ref{tab:my_table} presents the results of experiments conducted for $X_T\sim\mathcal{N}(0.3,0.5)$, $X_G\sim\mathcal{N}(0,1)$ and $X_D=X_T+\sum_{i=1}^4 Z_i$, $X_S=X_G+\sum_{i=1}^4 Z_i$ where $Z_i\sim\mathcal{N}(0,1)$.  There are no noise, efficiency, or acceptance effects.  We consider 5 scenarios where the deconvolution is performed using only $X_D$ and $X_S$, $X_D,X_S,Z_0$, ..., and $X_D,X_S,Z_0,...,Z_4$.  For the five-dimensional case, there are actually no resolution effects since $X_T$ can be uniquely determined from $X_D$ and the $Z_i$.  The table shows that more features leads to better accuracy and precision for fewer iterations of the algorithm.

By using classifiers instead of ratios of density estimators for the weighting steps, we can also make use of advances in classification networks to handle complex and structured data.  For example, deep sets~(\cite{zaheer2018deep,Komiske:2018cqr}) was used for a permutation-invariant and variable-length example from collider physics in Ref.~(\cite{Andreassen:2019cjw}), excluding background noise and acceptance/efficiency effects.

\begin{table}[]
    \centering
    \begin{tabular}{ccccc|ccccc}
     & \multicolumn{4}{c}{$\bar{x}$ mean ($\times 10^2$)} & \multicolumn{4}{c}{$\bar{x}$ standard deviation ($\times 10^3$)}  \\
    iterations $\rightarrow$ & 1 & 2 & 4 & 8 & 1 & 2 & 4 & 8\\
    $N$   &  &&&&&&&\\ 
    \hline
1 & 21.62(8)& 25.13(8)& 28.12(8)& 29.67(8) & 8.4(5)& 8.3(6)& 8.0(7)& 7.9(7) \\
2 & 28.54(5)& 29.24(6)& 29.88(6)& \textbf{30.06(5)} & 5.3(4)& 5.6(4)& 5.2(4)& 4.8(4) \\
3 & 29.54(4)& \textbf{29.91(4)}& \textbf{30.02(4)}& \textbf{30.00(4)} & 3.6(3)& 4.4(4)& 3.6(3)& 4.0(3) \\
4 & 29.89(3)& \textbf{30.01(3)}& \textbf{30.01(3)}& \textbf{30.01(3)} & \textbf{3.2(2)}& \textbf{2.8(2)}& \textbf{3.1(2)}& \textbf{3.1(2)} \\
5 & \textbf{30.04(3)}& \textbf{30.00(3)}& \textbf{29.99(4)}& \textbf{30.06(3)} & 3.5(3)& \textbf{3.1(2)}& 3.8(3)& \textbf{3.1(2)} \\
    \end{tabular}
    \caption{The average and standard deviation of the mean $\bar{x}$ of the unfolded $\hat{X}_T$ in the multidimensional Gaussian example.  Uncertainties are determined over 100 identical experiments.  Bolded values are within $2\sigma$ of the correct value (left) or are consistent with the smallest uncertainty (right).}
    \label{tab:my_table}
\end{table}

\section{Conclusions and Outlook}
\label{sec:conclusions}

In this paper, we have introduced the \textsc{OmniFold} method, which scaffolds a forward simulation with deep learning in order to perform unbinned and high-dimensional deconvolution.  We have extended the framework from Ref.~(\cite{Andreassen:2019cjw}) to include the full suite of experimental effects that are encountered in practice and we have demonstrated this approach using a Gaussian example.  Our code is available at \url{https://github.com/hep-lbdl/OmniFold} and can be deployed for a wide variety of scientific applications ranging from differential cross section measurements in high energy physics to quantum computing~(\cite{1910.01969}).

%\subsubsection*{Author Contributions}
%If you'd like to, you may include  a section for author contributions as is done
%in many journals. This is optional and at the discretion of the authors.

\subsubsection*{Acknowledgments}
BN and AS are supported by the U.S. Department of Energy (DOE), Office of Science under contract DE-AC02-05CH11231.  BN thanks Miguel Arratia for many useful discussions and helpful feedback about the code.
JT and PK are supported by the National Science Foundation under Cooperative Agreement PHY-2019786 (The NSF AI Institute for Artificial Intelligence and Fundamental Interactions, \url{http://iaifi.org/}), and by the U.S. DOE Office of High Energy Physics under grant number DE-SC0012567.

%\clearpage

\bibliography{iclr2021_conference}
\bibliographystyle{iclr2021_conference}

%\appendix
%\section{Appendix}
%You may include other additional sections here.

\end{document}